\documentclass[letterpaper, 10 pt, conference]{ieeeconf}  

\IEEEoverridecommandlockouts                              

\usepackage[utf8]{inputenc} 
\usepackage[T1]{fontenc}    
\usepackage{hyperref}       
\usepackage{url}            
\usepackage{booktabs}       
\usepackage{amsfonts}       
\usepackage{nicefrac}       
\usepackage{microtype}      

\usepackage{amsmath}
\usepackage{amssymb}
\usepackage{xspace}
\usepackage{dsfont}
\usepackage{graphicx}
\usepackage{caption}
\usepackage{subcaption}
\usepackage{adjustbox} 
\usepackage{multirow} 

\newcommand{\be}{\begin{equation}}
\newcommand{\ee}{\end{equation}}

\newcommand*{\ie}{i.e.\@\xspace}
\newcommand*{\iid}{i.i.d.\@\xspace}
\newcommand*{\resp}{resp.\@\xspace}
\newcommand\norm[1]{\left\lVert#1\right\rVert}

\DeclareMathOperator*{\argmax}{arg\,max}

\title{\LARGE \bf Dropout Distillation for Efficiently Estimating Model Confidence}

\author{Corina Gurau$ $, Alex Bewley and Ingmar Posner
\thanks{$ $Corina Gurau, Alex Bewley and Ingmar Posner are with the Oxford Robotics Institute, United Kingdom
        {\tt\small corina@robots.ox.ac.uk, alex.bewley@gmail.com, ingmar@robots.ox.ac.uk}}%
}

\begin{document}

\maketitle
\thispagestyle{empty}
\pagestyle{empty}

\begin{abstract}
We propose an efficient way to output better calibrated uncertainty scores from neural networks.
The Distilled Dropout Network (DDN) makes standard (non-Bayesian) neural networks more introspective by adding a new training loss which prevents them from being overconfident.
Our method is more efficient than Bayesian neural networks or model ensembles which, despite providing more reliable uncertainty scores, are more cumbersome to train and slower to test.
We evaluate DDN on the the task of image classification on the CIFAR-10 dataset and show that our calibration results are competitive even when compared to 100 Monte Carlo samples from a dropout network while they also increase the classification accuracy.
We also propose better calibration within the state of the art Faster R-CNN object detection framework and show, using the COCO dataset, that DDN helps train better calibrated object detectors.
\end{abstract}


\section{Introduction}

Deep neural networks are now the default model choice due to their impressive performance on a wide range of tasks and benchmarks.
Predictive uncertainty estimates from standard neural networks are typically overconfident, often making them too unreliable to be deployed in real world applications.
Bayesian Deep Learning provides uncertainty estimates at the cost of posing neural networks as approximate Bayesian models and drawing multiple posterior samples at test time 
The timing of this approach scales linearly with the number of samples, making Bayesian neural networks often too slow to use in practice.
In this work we propose a method for improving the uncertainty estimates from neural networks that is simple to implement and more readily suitable to use in real-time applications.

By treating the model parameters as a random variable, Bayesian probability theory allows us to compute uncertainty in model predictions.
Given a dataset $\mathcal{D}=\{\mathcal X, \mathcal Y\}$ with output prediction $\mathbf{y} \in \mathcal Y$ for a new input $\mathbf{x} \in \mathcal X$, we can compute the posterior predictive distribution:
\be
    p(\mathbf{y}|\mathbf{x}, \mathcal D) = \int p(\mathbf{y}| \mathbf{x}, \mathbf{w}) p(\mathbf{w}|\mathcal D)dw, \label{eq:posterior}
\ee
by using the likelihood $p(\mathbf{y}|\mathbf{x}, \mathbf{w})$ and a posterior distribution over model parameters $p(\mathbf{w}|\mathcal D)$ which accounts for predictive uncertainty.
For a non-linear neural network however, the term $p(\mathbf{w}|\mathcal D)$ is analytically intractable.
Various approximations to it have been proposed, but Bayesian Neural Networks are known to be cumbersome to train and computationally expensive to use in practice (\cite{Blundell_weight_uncert}).

Dropout, a technique introduced by \cite{dropout} to improve model generalisation), has been shown by \cite{Gal2015DropoutB} to produce a Monte Carlo approximation to Equation~\ref{eq:posterior} and that by sampling this approximation one can obtain samples from the posterior predictive distribution.
These samples, referred to as MC-dropout samples, are often interpreted as an ensemble of models.
The downside of the approach proposed by \cite{Gal2015DropoutB} is that Monte Carlo sampling causes a linear increase in either time or resources, often making it difficult to use in real-time operation.
This issue is acknowledged in~\cite{Gal2016Uncertainty} as well as in~\cite{simple_and_scalable}.

We propose a practical alternative for producing reliable uncertainty estimates at test time more efficiently, without the cumbersome sampling procedure.
We propose to learn, in a student-teacher paradigm, to produce accurate uncertainty values by observing the Monte Carlo dropout samples at training time only.
In essence, the framework we propose amounts to training two models: one that is simply trained to maximise task accuracy (the \textit{teacher}), and one that is specifically trained to output better uncertainties (the \textit{student}).
We refer to the latter as the \textit{Distilled Dropout Network} (DDN), as it distills knowledge from multiple Monte Carlo samples from the teacher in order to improve the reliability of its uncertainty scores.

Our experiments empirically demonstrate that the uncertainty scores produced by DDN are indeed comparable to those provided by an ensemble of MC-dropout predictions.
The proposed model is also computationally efficient as it only requires a single forward-pass and does not rely on sampling at test time.
Additionally, we show that DDN also increases the task accuracy, although that is not our main objective.

\section{Related Work}
Model ensembles are known to significantly improve predictive performance (\cite{ensemble_methods_in_ml}) as well as to provide a way to measure uncertainty in the space of predictors.
Methods for producing ensembles such as boosting, bagging (bootstrap aggregating) or stacking rely on fitting multiple models to subsets of the training data and averaging their predictions at test time.
This process reduces the variance of a single prediction and, according to \cite{GrimmettIJRR2015}, leads to an increased \textit{introspective} capacity.
Introspection refers to the ability of a model to associate an appropriate assessment of confidence to its predictions.

In a similar manner to \cite{regularising_hinton} and \cite{BucilaCN06}, we train a student network to match the probabilities that a high capacity (teacher) network assigns to all output classes, a process referred to as distillation.
Unlike \cite{regularising_hinton} and \cite{BucilaCN06}, we do not seek a more compact version of the teacher model, but rather a better calibrated version of it.
Rather than reducing the capacity of the student, we keep its architecture identical to the teacher's in a similar way to \cite{BANNs}, who call such models Born Again Networks (BANs).

The name \textit{dropout distillation} first appeared in \cite{Bulo_dropout_distill} as an attempt to train a model that minimises the divergence between itself and an ideal predictor (obtained by averaging over multiple Monte Carlo samples) at training time.
Their work is different from ours as they only seek to improve accuracy and do not address model calibration. 

Our work is also similar to \cite{simple_and_scalable} who seek better calibrated models and propose the use of deep ensembles.
They show that a non-Bayesian solution (ensembles) can perform just as well as Bayesian inference in producing predictive uncertainty.
Their research findings were encouraging in the direction of this work.
However, they do not address the memory or computational increase introduced by model ensembles.
Our research shows that an approximation to an ensemble is still effective at producing reliable uncertainty scores.

In \cite{on_calibr_on_nn} the authors propose to analyse some factors that cause the poor calibration of deep networks and identify that the loss choice, network architecture and weight decay all have an influence in this phenomenon.
Our work is similar to theirs particularly in the evaluation procedure, as they measure calibration, accuracy and risk of neural networks used for image classification.

Penalising confident (low entropy) output distributions is seen as a form of regularisation by \cite{regularising_hinton}.
They use an entropy-based loss term in order to penalises peaked distributions rather than using distillation or self-distillation in order to improve class confidence.
While we do not address model regularisation, their loss function shares similarities with ours.

\section{Approach}
Our idea is inspired by \cite{distilling}, which transfers knowledge from an ensemble of models to a single (smaller) model, a process for which they coin the term distillation.
Our goal differs from that of \cite{distilling} in the sense that we do not seek a more compressed version of the ensemble (the teacher) model but rather a better calibrated version of it.
The teacher's prediction is the mean of the multiple dropout samples.

\subsection{MC-Dropout} \label{sec:mc-dropout}
The idea of dropout was introduced by \cite{dropout} as a way to regularise a neural network by allowing the units to be masked during training in order to prevent their co-adaptation and therefore limit overfitting. 
The mask on each weight is drawn from a Bernoulli distribution $Ber(\theta)$, where $\theta$ controls the probability of not dropping that weight.
\cite{dropout} interpret dropout as producing an ensemble of $2^n$ models, where $n$ is the number of units in a single model.
\cite{Gal2015DropoutB} as well as \cite{kendall2017uncertainties} have shown that the stochasticity introduced by dropout allows the network to sample the model space.

Assuming that $f^{\theta}_w$ is a network parametrised by $\mathbf{w}$ with a dropout mask generated by $Ber(\theta)$, sampling it:
\be
\widehat{\mathbf{y}} \sim f^{\theta}_w(\mathbf{x}),
\ee
allows us to obtain multiple output samples $\widehat{\mathbf{y}}_i$ from stochastic forward-passes through the network.
The predictive mean of the ensemble is the average of the $M$ output samples,
\be
\mathbb{E}_{f_w^{\theta}}(\mathbf{y}) \approx \dfrac{1}{M} \displaystyle \sum_{i=1}^{M} \widehat{\mathbf{y}}_i,
\ee
and the predictive variance is:
\be
\mathrm{Var}_{f_w^{\theta}}(\mathbf{y})\approx \dfrac{1}{M} \displaystyle \sum_{i=1}^{M} \widehat{\mathbf{y}_i}^\top\widehat{\mathbf{y}_i}-\mathbb{E}(\mathbf{y})^\top\mathbb{E}(\mathbf{y}).
\ee
With infinite, non-\iid training data, this variance converges to $0$, but given the bias currently present in many computer vision datasets, models learnt on such datasets are far from having a low variance when tested in real-world robotics applications.

\subsection{Knowledge Distillation} \label{sec:distill}

One of the most commonly used losses for training neural networks that output a categorical distribution is the cross entropy loss: 
\be
    \mathcal{L}(\mathbf{w}) = -\sum_{i=1}^{N} \mathbf{y}_i^T\log{(\sigma(f_w(\mathbf{x}_i)))},  \label{eq:xent_1}
\ee
where $\mathbf{y}_i$ is a one-hot encoding of the ground truth labels ($1$ corresponding to the true class and $0$ to all others) associated to sample $\mathbf{x}_i$.
The softmax function $\sigma$ ensures a probabilistic output of the vector $\mathbf{z} = f_w(\mathbf{x})$ of predicted logits.

With cross entropy, even when all predictions $f_w(\mathbf{x})$ are correct, the network can still decrease the training loss by increasing the probability assigned to the true class and decreasing the ones assigned to the rest of the classes.
We hypothesise that this is what usually leads to the overconfidence of state of the art neural networks.
Another problem with the cross entropy loss is that it encourages the assignment of high probability to a single class when in reality the model could be moderately confident in two or more classes (for instance, it is understandable to confuse cats and dogs, particularly when they are partially occluded).

Knowledge distillation, while having a different goal to ours, is also motivated by allowing networks to express this kind of uncertainty at training time.
In \cite{distilling}, the authors claim that seeing the probabilities assigned to incorrect classes as well as the correct class makes student models generalise better.
In order to obtain such targets (often called `soft targets'), their approach is to increase a temperature parameter in the teacher model's softmax function until it produces a softer distribution over the output classes.
They propose to use two losses: a `hard' cross entropy loss with the ground truth labels (Equation \ref{eq:xent_1}) as well as a `soft' cross entropy loss with a high temperature in the softmax function:
\be
    \sigma_i(\mathbf{z}) = \frac{e^{z_i/T}}{\sum_{j=1}^{C}e^{z_j/T}},   \label{eq:high_temp_softmax},
\ee
where $T$ a temperature parameter controlling how much to affect the output distribution.
In this work, we propose a distillation process that uses a similar mechanism in order to produce better calibrated uncertainty scores.
Rather than raising the temperature, we use the mean prediction of multiple dropout samples.

\subsection{The Distilled Dropout Network}

The Distilled Dropout Network (DDN) is trained to mimic the predictions of an ensemble, using distillation in such a way that it trains a better calibrated model.
Similar to \cite{distilling} we use both the one-hot ground truth labels and a proposed soft target distribution obtained by averaging the Monte Carlo samples.
However, the way we transfer knowledge is more similar to \cite{NIPS2014_5484}, as we use the logits $z$ as soft targets rather than the probabilities produced by the softmax.

Given training data $\mathcal{D} = \{(\mathbf{x}_i, \mathbf{y}_i, \mathbf{z}_i)\}_{i=1}^N$ containing a set of input examples $\mathbf{x}$, ground truth labels $\mathbf{y}$ and the corresponding set of distilled labels $\mathbf{z}$ obtained using a teacher model $g_{\mathbf{w^*}}^{\theta}$, we train the student model $f_\mathbf{w}$ by minimising the loss:
\be
    \mathcal{L}(\mathbf{w}) = \mathcal{L}_{\text{DD}}(\mathbf{w}) + \lambda \mathcal{L}_{\text{GT}}(\mathbf{w}) + \gamma \mathcal{L}_{\mathcal{R}}(\mathbf{w}). \label{ddn_loss}
\ee
The distilled dropout loss $\mathcal{L}_{\text{DD}}$ is defined as the $L_2$ loss with the teacher logits $z$:
\begin{equation}
    \mathcal{L_{\text{DD}}}(\mathbf{w}) = \dfrac{1}{N}\sum_{i=1}^{N}\norm{f_w(\mathbf{x}_i)-\mathbf{z}_i}^2_2,
\end{equation}
$\mathcal{L}_{\text{GT}}$ is the standard cross entropy loss with the ground truth labels $y$:
\begin{equation}
    \mathcal{L}_{\text{GT}}(\mathbf{w}) = -\dfrac{1}{N}\sum_{i=1}^{N} \mathbf{y}_i\log({\sigma{(f_w(\mathbf{x}_i)}))},
\end{equation}
and $\mathcal{L}_{\mathcal{R}}$ is a regularisation loss.
The hyperparameters $\lambda$ and $\gamma$ control the importance of the additional losses.

To obtain the logit target labels $\mathbf{z}$, we average $M$ MC-dropout predictions from the teacher model $g_{w^*}^{\theta}$ on exactly the same input data $\mathbf{x}$:
\begin{equation}
    \mathbf{z}_i = \frac{1}{M}\sum_{i=1}^{M}g_{w^*}^{\theta}(\mathbf{x}_i).
\end{equation}
Using additional data in order to generate more training samples for the student should also be possible.
This includes even samples without ground truth annotations which could be incorporated into the student training procedure without computing $\mathcal{L}_{\text{GT}}$. 
This idea will be the subject of future research.

\subsection{Accuracy, Calibration and Risk}
In this work, two (orthogonal) metrics are of interest: \textit{accuracy} and \textit{calibration}.
In a classification setting where the output vector $f_w(\mathbf{x})$ is a categorical distribution over $C$ output classes (where $f_w^c$ indicates the probability assigned to class $c$), accuracy is concerned with the precision of the model's class predictions $\hat y$, where
\be
\hat y = \argmax_{c}{f_w^c(\mathbf{x})}
\ee
while calibration is concerned with the quality of the probability value assigned to that prediction:
\be
f_w^{\hat y}(\mathbf{x}) = \max_{c}({f_w^c(\mathbf{x})}).
\ee
For instance, an introspective classifier assigning a probability value of $0.8$ to its predictions is expected to be correct roughly $8$ out of $10$ times.

A frequentist approach to measuring calibration is by computing the difference between the confidence predicted by a model and the empirical frequencies of successful classifications made by that model.
This empirical calibration metric has been introduced in \cite{degroot} and is a common choice for measuring the quality of predictive uncertainty used in works such as \cite{kendall2017uncertainties}, \cite{on_calibr_on_nn}, \cite{Gal2017Concrete} and \cite{OndruskaIVS2014}.
We measure calibration in a similar way to \cite{on_calibr_on_nn}, by assigning all output probabilities $f_w^{\hat y}(\mathbf{x})$ corresponding to the predicted classes $\hat y$ to $K$ equally sized bins such that $D_k$ is the set of predictions for which $f_w^{\hat y}(\mathbf{x}) \in \big (\frac{k-1}{K}, \frac{k}{K}\big ]$, and computing the success frequency of the predictions in bin $D_k$:
\be
\text{freq}(D_k) = \frac{1}{\left\vert D_k \right\vert}\sum_{\mathbf{x}_i\in D_k}\mathds{1}(\hat y_i = y_i),
\ee
as well as the confidence of the predictions in bin $D_k$:
\be
\text{conf}(D_k) = \frac{1}{\left\vert D_k \right\vert}\sum_{\mathbf{x}_i\in D_k} f_w^{\hat y}(\mathbf{x}).
\ee

Our calibration metric of interest is the Mean Squared Calibration Error (MSCE) computed as:
\be
\text{MSCE} = \sum_{i=1}^{K} \frac{\left\vert D_i \right\vert}{N} {\lvert \text{freq}(D_i) - \text{conf}(D_i) \rvert}^2.
\ee
Maximum Calibration Error (MCE) is similar to MSCE but instead of the mean it computes the maximum error between confidence and success frequency. 
In the experimental evaluation we show these metrics using calibration plots, where a perfect calibration corresponds to a straight line from (0, 0) to (1, 1).
Deviation from the diagonal towards the bottom right of the plot indicates overconfidence while deviation towards the top left indicates underconfidence in the predictions.
The lower the MSCE and MCE scores are, the better calibrated the model is.

Two Bayesian measures of uncertainty are the normalised negative log-likelihood ($\mathrm{NLL}$): 
\be
    \mathrm{NLL}(\mathbf{w}) = -\frac{1}{N}\sum_{i=1}^{N}\log f_w^{\hat y}(\mathbf{x}_i), \label{eq:nll}
\ee
and the entropy $H$ of the prediction:
\be
    H(\mathbf{w}) = -\frac{1}{N}\sum_{i=1}^{N}\sum_{j=1}^{C} f_w^j(\mathbf{x}_i) \log f_w^j(\mathbf{x}_i),
    \label{eq:xent}
\ee
where $C$ is the number of classes and $N$ is the number of examples. 
We empirically show that these measures are not indicative of how well calibrated a model is.


\section{Classification Experiments}

We first evaluate the outcome of our proposed training procedure on the task of image classification, and then extend its usage to the problem of object detection in Section \ref{sec:frcnn_res}.

\subsection{Dataset and Network Architecture}

We use the CIFAR-10 dataset (\cite{Krizhevsky09learningmultiple}) for image classification.
The dataset consists of coloured natural images of size $\mathrm{32 \times 32}$ pixels drawn from $10$ different classes.
The training and test sets contain 50k and 10k images respectively, with an equal class balance.

As a baseline we train a Dense Convolutional Network (DenseNet) model architecture introduced in \cite{huang2017densely}. 
We use the $40$-layer configuration with a growth rate of 12 described in \cite{huang2017densely}.
More precisely, we train the network using stochastic gradient descent (SGD) using batches of size $64$ for $300$ epochs. 
The initial learning rate is set to $0.1$, and is then divided by 10 at $50\%$ and $75\%$ of the total number of training epochs.
We use a weight decay of $10^{-4}$ and a Nesterov momentum of $0.9$ (\cite{pmlr-v28-sutskever13}).
The dropout rate (\ie the probability of retaining a unit in the network) has been set to $0.8$.
All hyperparameters mentioned have been chosen according to the original DenseNet work (\cite{huang2017densely}).

We use this model as the teacher and generate $100$ Monte Carlo samples per input for the knowledge distillation process.
The dropout rate for generating these targets has also been set to $0.8$ (typically using a dropout rate different than the one used in training causes less accurate predictions as shown in \cite{Gal2017Concrete}).
DDN is trained from scratch using an identical network architecture, training regime and hyperparameter values, the only difference being the loss function described in Equation~\ref{ddn_loss}.

At test time, both networks have a dropout rate set to $1$.
We occasionally refer to the baseline model as `No Dropout' to indicate a standard single forward-pass through the network with no dropout.
Although they have the same weights, the distinction between the baseline and the teacher is that the latter is stochastic (dropout rate set to $0.8$) in order to generate multiple Monte Carlo samples.

\subsection{Results}

\begin{figure}[]
    \centering
    \includegraphics[width=0.4\textwidth]{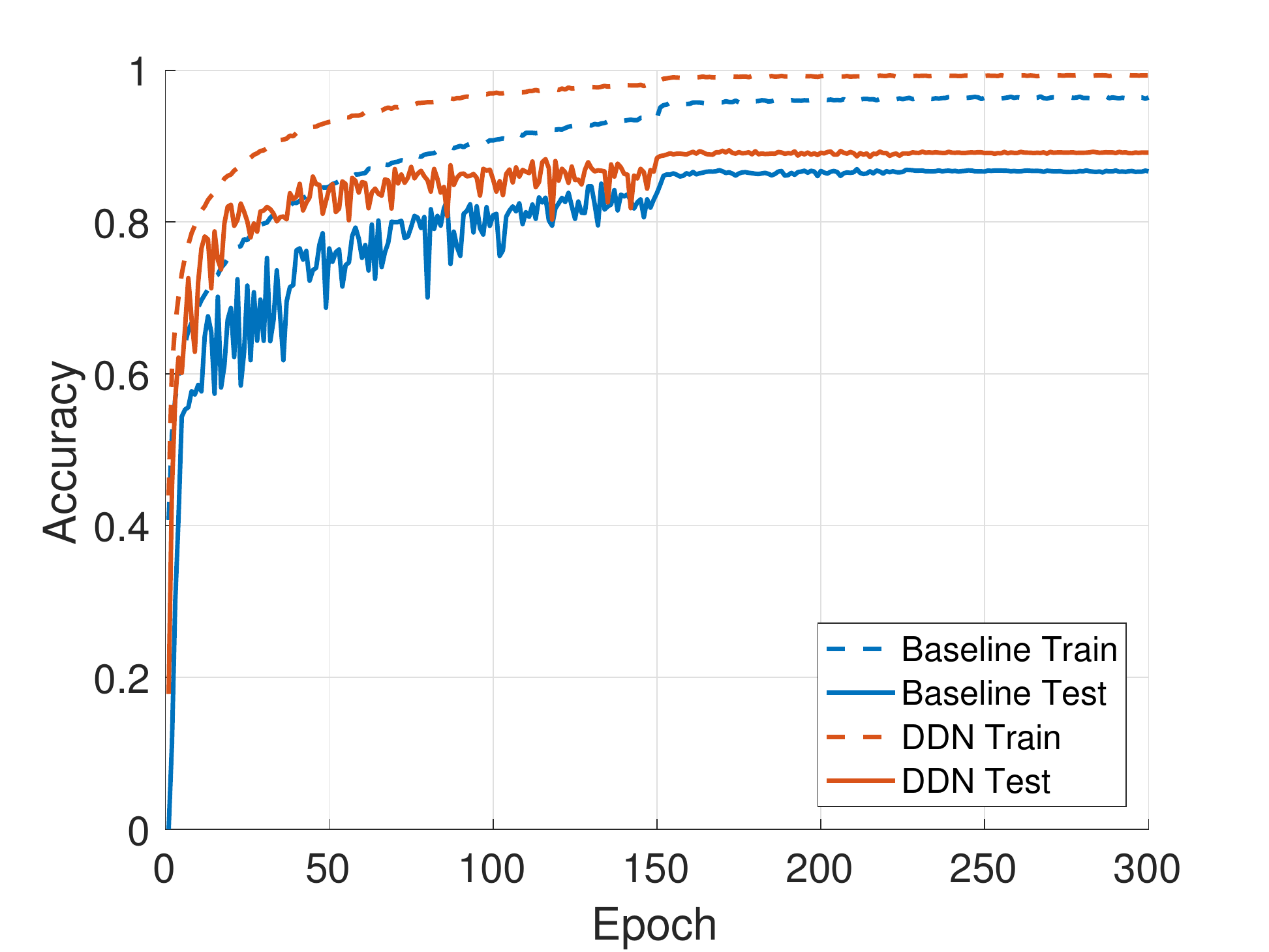}
    \caption[CIFAR-10 training and testing accuracy]{Figure showing the accuracy of the proposed student model (DDN) as well as a baseline teacher model during training and testing on the CIFAR-10 dataset. DDN achieves better test accuracy than the teacher model, despite being trained on the same data with an identical training regime.}
    \label{fig:accuracy}
\end{figure}
\begin{figure}[]
    \centering
    \includegraphics[width=0.4\textwidth]{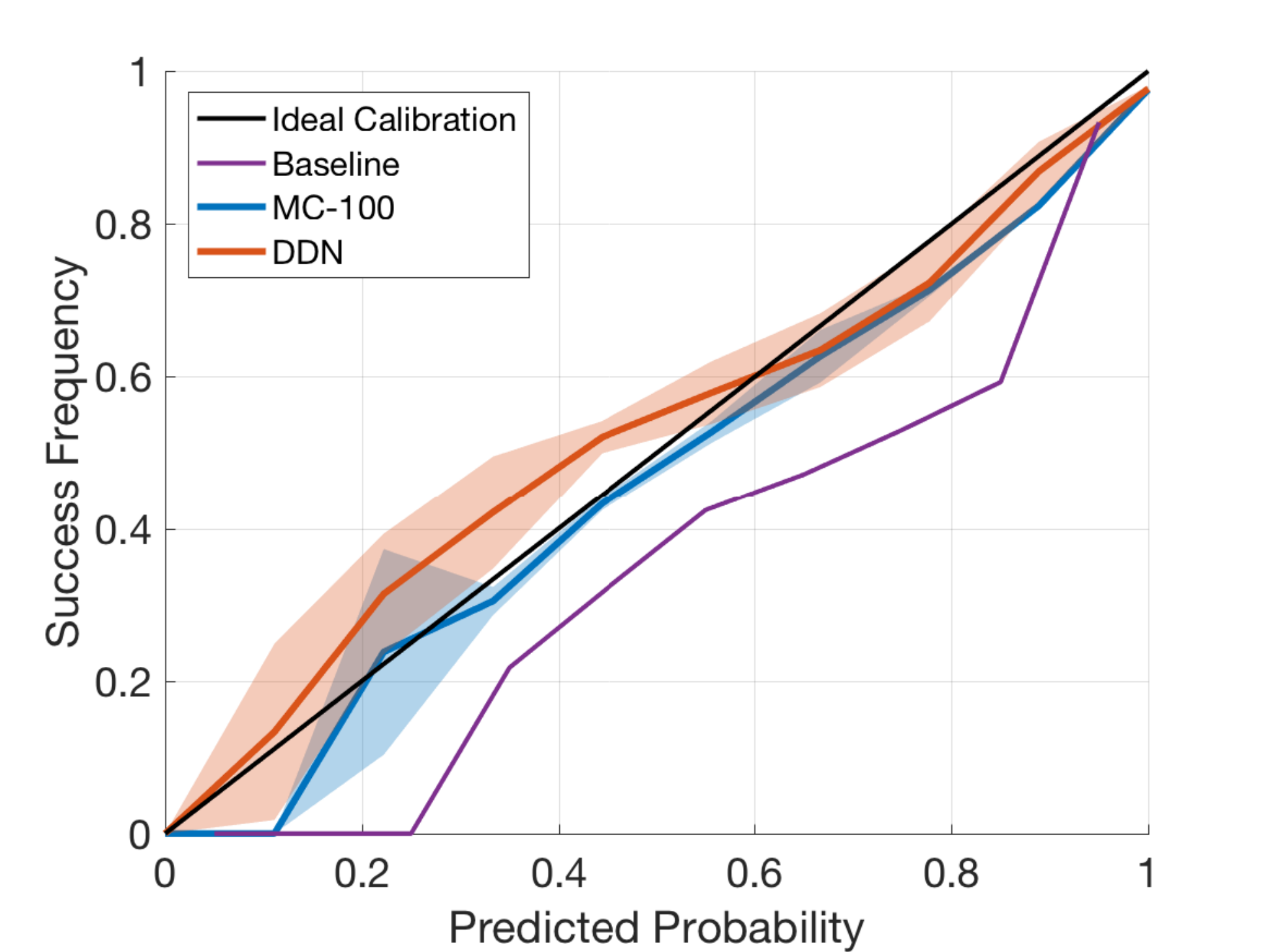}
    \caption[Calibration plot on CIFAR-10 test data]{Calibration plot showing the correlation between class uncertainty and (empirical) success frequency. DDN is our proposed student model that can produce high quality uncertainty scores in a single forward-pass through the network. The baseline model corresponds to using all the units in the teacher model and MC-100 is the outcome of sampling the model 100 times and averaging the predictions.}
    \label{fig:calibration}
\end{figure}  

\begin{figure}[]
    \centering
    \includegraphics[width=0.4\textwidth]{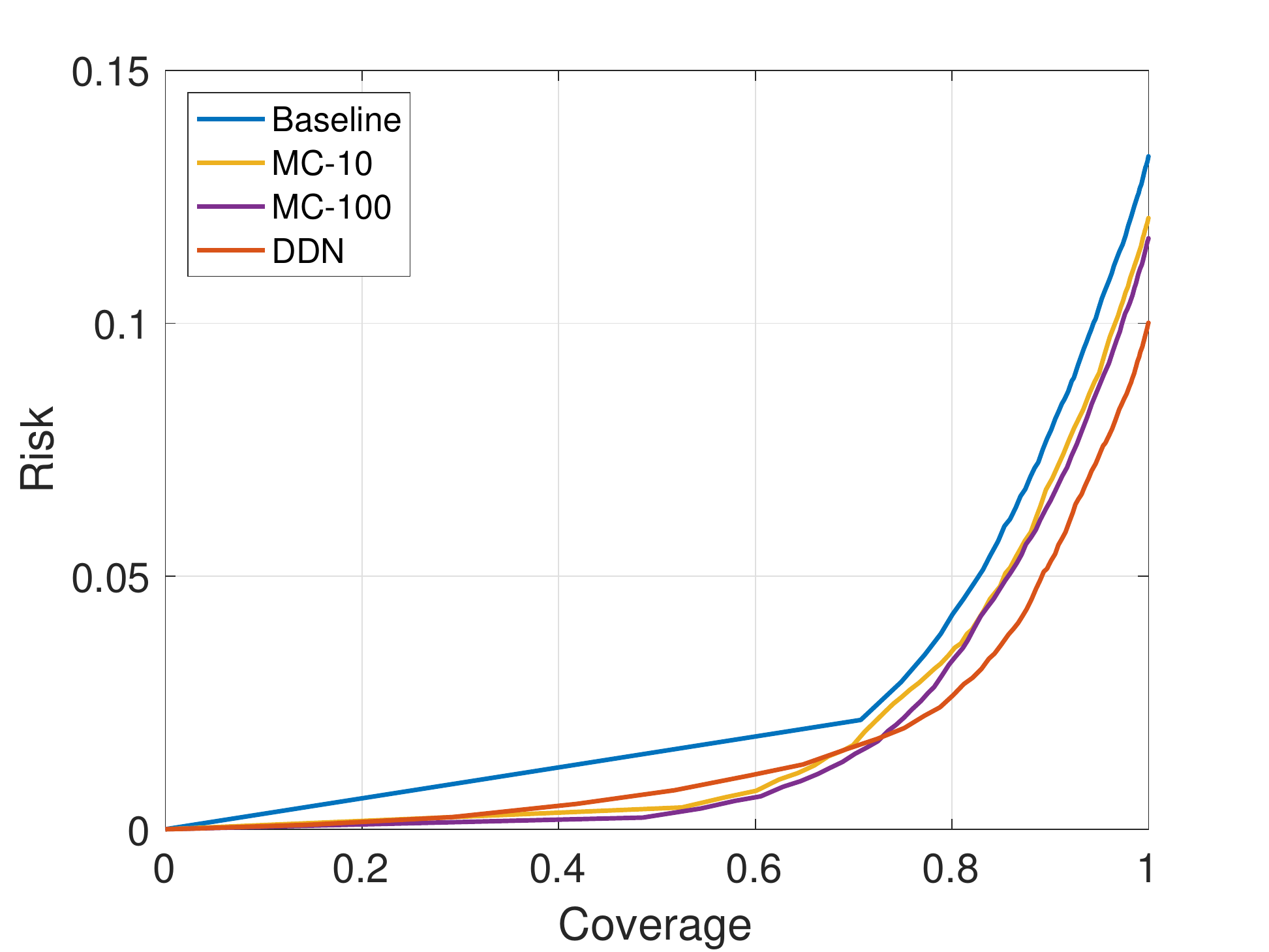}
    \caption{Risk-Coverage plot showing the percentage of mistakes (risk) for a given coverage of the test data. The predictions are ordered by confidence such that in the low coverage areas of the graph only the most confident predictions are used. A coverage value of $1$ corresponds to using all the samples in the test set.}
    \label{fig:risk}
\end{figure}

DDN achieves better training and testing accuracy than the baseline model as shown in Figure~\ref{fig:accuracy}.
The final test accuracy of the teacher is $87\%$, while the student model can reach $90\%$ test accuracy.
This is also confirmed by the works of \cite{regularising_hinton} and \cite{distilling}.
The former claims that encouraging low-entropy predictions is a form of model regularisation which helps the model's generalisation capabilities.
The latter attributes it to the fact that the student model has knowledge of the ratios between output class confidence scores, while the teacher does not.

As expected, the student model produces higher entropy output probabilities than the baseline model.
Of the highest scoring predictions from the DDN, only $38\%$ have a confidence score above $0.95$ compared to $77\%$ of the predictions made using the baseline model with no dropout.
This indicates that the baseline is not only less accurate in its predictions, but also more confident.
DDN does not explicitly penalise low-entropy (overconfident) distributions in training but rather encourages predictions to also assign some probability mass to classes which it is less sure about.

Figure~\ref{fig:calibration} shows that the probability scores assigned to the test samples by DDN are better calibrated (closer to the ideal calibration) than those assigned by the baseline model.
The training uses as targets $100$ MC-dropout samples from the teacher model.
Throughout this work the teacher has the same model weights as the baseline, the only difference being that its dropout rate has been set to $0.8$ instead of $1$ in order to produce stochastic samples.
This figure also confirms that the DDN predictions are slightly under-confident, while the baseline model predictions are overconfident, a bias that tends to be preferred in high-risk applications.
We also compare the baseline and DDN models to a very well calibrated model given by $100$ Monte Carlo samples (MC-100).
The variance in the distilled models as well as MC-100 is a result of the stochasticity in the training procedure (dropout, random shuffling of training samples) observed in retraining the models.
The plot shows the mean and variance of three distilled models and three MC-100 models.

The model's risk, shown in Figure~\ref{fig:risk}, is defined as as the percentage of mistakes made for a particular coverage of the dataset (\ie the proportion of test samples used).
The coverage is computed as the percentage of test samples where the model's confidence is above a threshold which we vary from $0$ to $1$ in Figure~\ref{fig:risk}.
The figure shows that at a low coverage of the dataset, a low risk is expected since only the most confident predictions are used.
However, for a higher coverage, DDN has the lowest risk, followed by MC-100.
The risk values in Table~\ref{tab:results_cifar_1} correspond to a full coverage of the test set.

Table~\ref{tab:results_cifar_1} shows the calibration scores for the four models with DDN being a competitive alternative to MC-100.
Besides the quality of the calibration, an important distinction between the models is the time it takes to produce a prediction.
As this metric depends on the model architecture, we measure it in the number of forward-passes through the network.
The ensemble models require, 10 \resp 100 forward-passes, while our baseline and DDN models make a single prediction.

Table~\ref{tab:results_cifar_1} also shows that the baseline model has the best NLL score, while the other calibration metrics indicate that it has the worst calibration.
Lower NLL and entropy values indicate increased confidence in the prediction but do not necessarily correlate well with the accuracy of the model.

\begin{table}
\centering
\begin{tabular}{ccccc}
\hline
Metric & \textbf{Baseline} & \textbf{MC-10} & \textbf{MC-100} & \textbf{DDN} \\ \hline\hline
MSCE & $0.9676$ & $0.0334$ & $\mathbf{0.0330}$ & \underline{$0.0927$} \\ \hline
MCE & $5.0694$ & \underline{$0.8898$} & $\mathbf{0.5250}$ & $1.0790$ \\ \hline
Risk & $0.1331$ & $0.1209$ & \underline{$0.1169$} & $\mathbf{0.1002}$ \\ \hline
NLL & $\mathbf{0.0634}$ & $0.1293$ & $0.1318$ & \underline{$0.1267$} \\ \hline
Entropy &  $\mathbf{0.1331}$ & \underline{$0.2661$} & $0.2790$ & $0.4445$ \\ \hline
Time & $\mathbf{1}$ & \underline{$10$} & $100$ & $\mathbf{1}$ \\ \hline
\end{tabular}
\caption[DDN calibration and timing results on CIFAR-10 test set]{
Calibration results on CIFAR-10 test set using various metrics. 
Best outcome is displayed in bold and second best is underlined.
The baseline model is a teacher model used with no dropout and a single forward-pass through the network.
MC-10 and MC-100 are ensemble models given by 10 and 100 stochastic samples from the baseline model.
DDN is our proposed student model learnt from scratch via (dropout) knowledge distillation from the baseline.
}
\label{tab:results_cifar_1}
\end{table}

\section{Object Detection Experiments} \label{sec:frcnn_res}

Our proposed dropout distillation method can also produce better calibrated object detector models.
We show this by using the Faster R-CNN framework acknowledged for its state of the art performance (\cite{ren2015faster}).
While making minimal modifications to the original framework, we fine-tune the original models with a distillation loss until calibration is improved.

\subsection{Dataset and System Architecture}

All experiments are performed on the COCO dataset (\cite{mscoco}).
The COCO dataset consists of roughly 80k training, 35k validation and 5k test images.
The dataset covers $81$ class categories which includes the `background' class usually interpreted as no detection of interest.
The network architecture considered is ResNet (\cite{He2015}).
Because a Faster R-CNN detection model requires significantly more time and computation to train from scratch, the student model is only fine-tuned using the distillation loss.
As teacher models (baselines) we use the original Faster R-CNN models trained on COCO: ResNet-50 and ResNet-101 trained with the open-source implementation of \cite{chen17implementation}.

Faster R-CNN is a two-stage detection framework, where the first stage consists of a region proposal network (RPN) that produces class-agnostic object proposals called Regions of Interest (RoIs).
The RPN is a fully convolutional network that simultaneously predicts object bounds and `objectness' scores at each position.
The second stage of the Faster R-CNN framework consists of a bounding box classifier as well as a bounding box regressor which take as input the RoIs and outputs the final bounding box scores and locations.
The two stages are trained jointly.

For our purpose, once a baseline model has been trained, we freeze the RPN model weights.
The distilled dropout loss will only improve the object classifier, while the RPN will produce the same set of proposal boxes as the original models.
We keep all classes and all image data exactly as in the original training procedure.
As is common practice, the models are pre-trained on the ImageNet dataset (\cite{imagenet}) and fine-tuned on the COCO training and validation data for 490k epochs.

For each of these baseline models, we generate the DDN target logits from multiple MC samples.
Due to computation constraints, we limit ourselves to $10$ samples.
We test different overlap threshold scores in order to generate the logit targets data ($0.3$, $0.5$ and $0.7$).
For the ground truth bounding boxes that do not have any detections associated to them from the sampling procedure, we set as target for the distillation loss a uniform distribution over the output classes to encourage the student model to be very uncertain.
This is motivated by the fact that since no detections were associated to it, the ground truth bounding box must be a difficult sample.
Incorporating this type of information in the training procedure helps prevent DDN from making mistakes with high confidence.

All hyperparameters are set to values recommended by \cite{chen17implementation}.
In training both the baseline models and the DDNs, we compute losses only on predictions that score higher than $0.05$, with a minibatch of $128$ examples per image and a single image per batch.
We use an initial learning rate of $0.001$ with a reduction factor of $0.1$ after $30000$ epochs, a weight decay of $0.0005$, a momentum of $0.9$ and an IoU overlap between a RoI and a ground-truth bounding box of $0.5$ used to compute $\mathcal{L}_{\text{GT}}$.
The number of training iterations to produce the baseline models is $490k$.

\begin{figure*}[]
\centering
    \begin{subfigure}[]{0.32\textwidth}
    \includegraphics[width=1\textwidth]{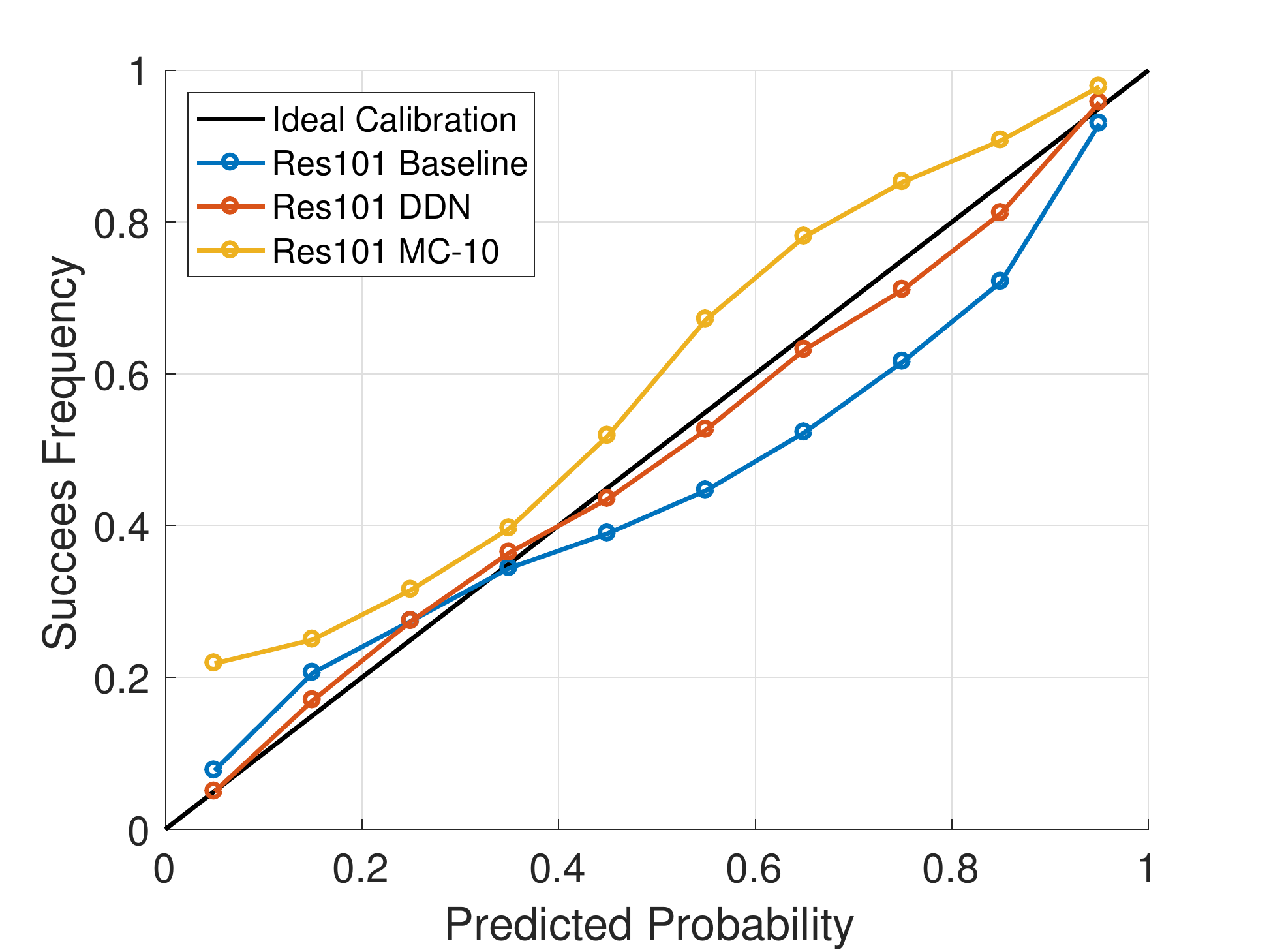}
    \caption{IoU threshold of $0.3$}
    \label{iou_03} 
\end{subfigure}
\begin{subfigure}[]{0.32\textwidth}
    \includegraphics[width=1\textwidth]{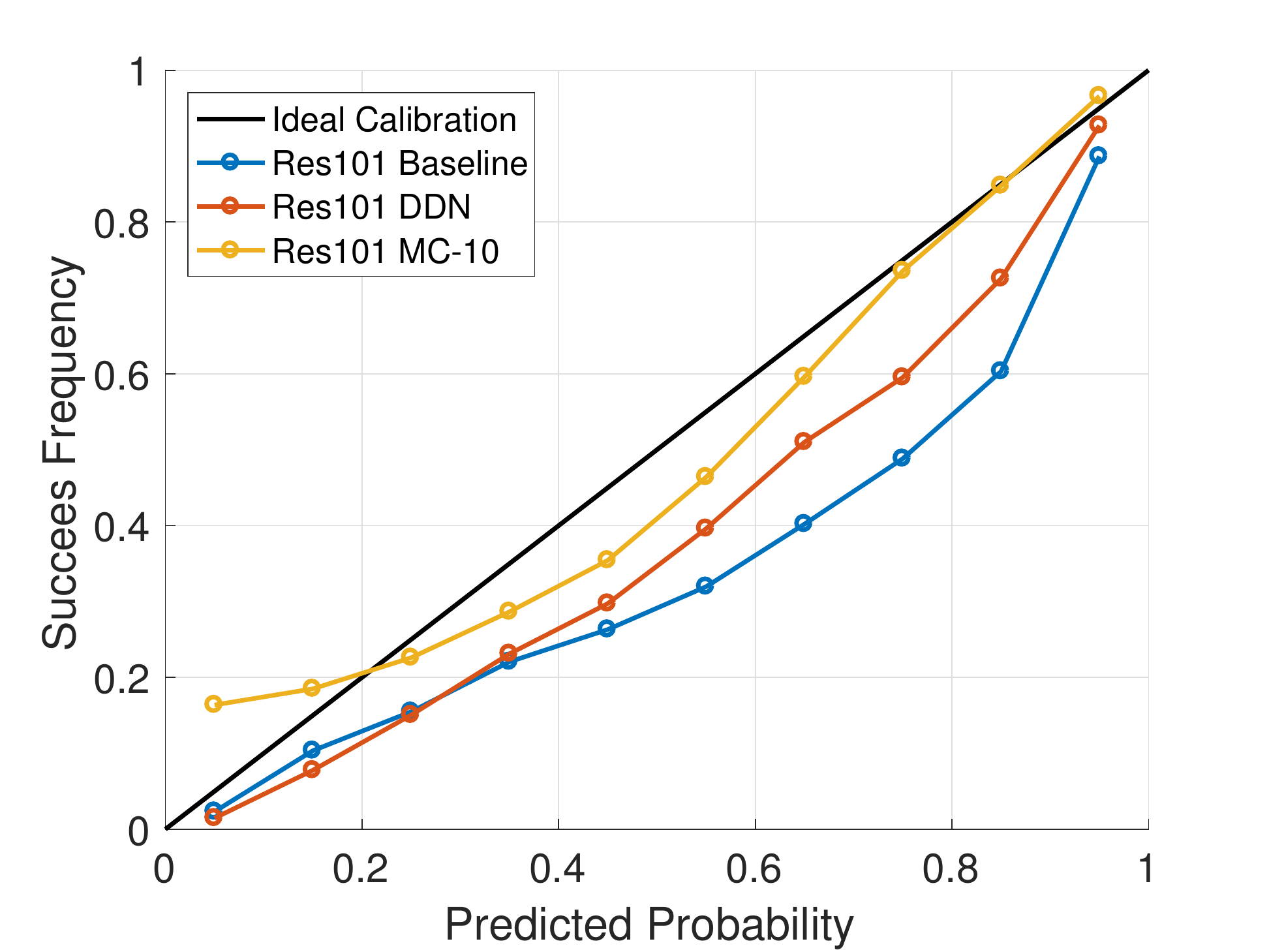}
    \caption{IoU threshold of $0.5$}
    \label{iou_05}
\end{subfigure}
\begin{subfigure}[]{0.32\textwidth}
    \includegraphics[width=1\textwidth]{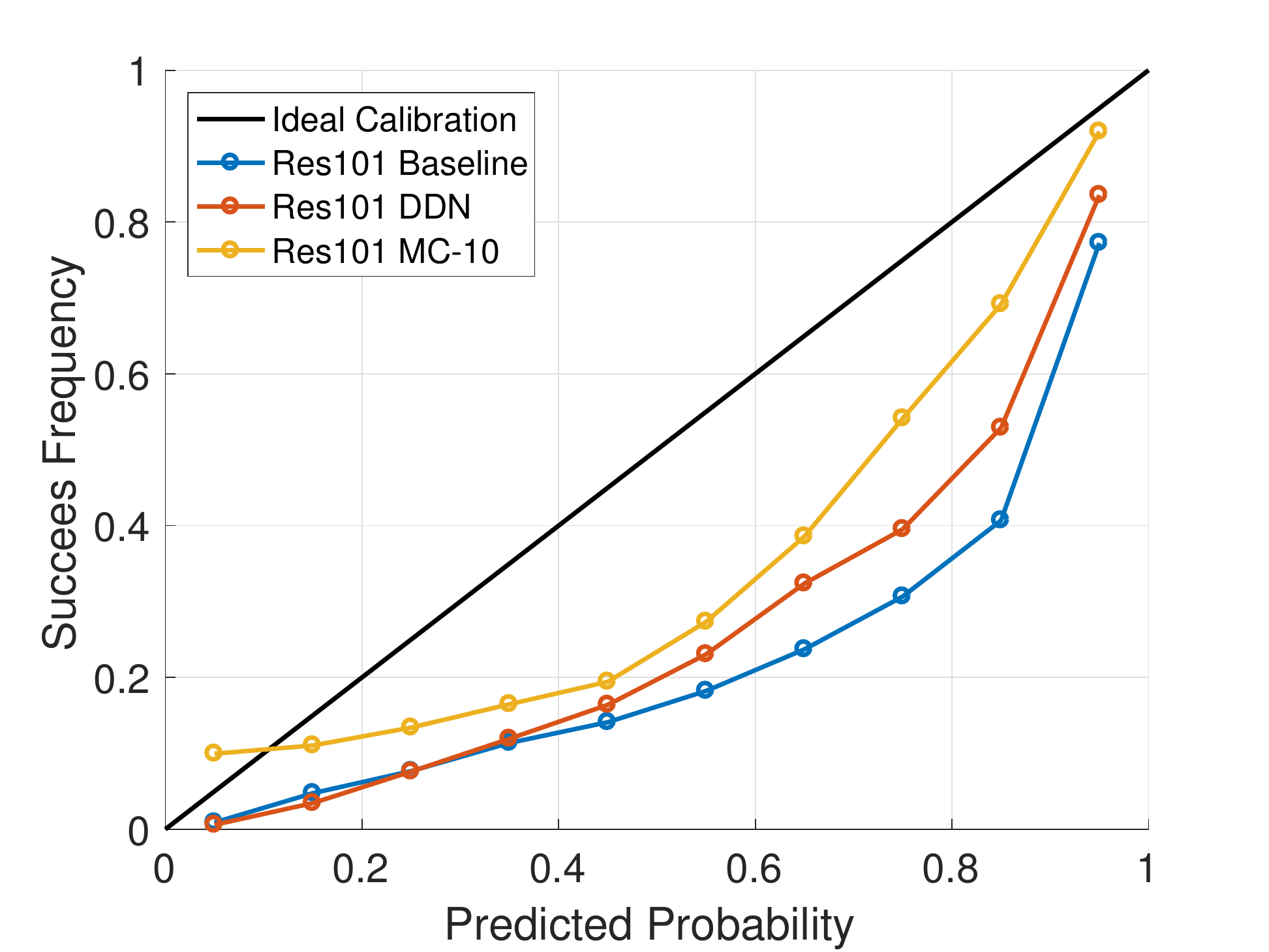}
    \caption{IoU threshold of $0.7$}
    \label{iou_07}
\end{subfigure}
\caption[Calibration plot on Faster R-CNN test data]{
Calibration plot showing the correlation between class uncertainty and (empirical) success frequency of the baseline ResNet50 and ResNet101 models as well as DDN. 
The three subfigures correspond to different IoU overlap thresholds used to distinguish between true and false positive detections.
Note that this IoU threshold is different than the one used for obtaining the target scores for distillation which has been set to $0.5$.
}
\label{fig:frcnn_calib_res}
\end{figure*}

\subsection{Results}

\begin{figure*}[]
\centering
    \begin{subfigure}[]{0.28\textwidth}
    \includegraphics[width=1\textwidth]{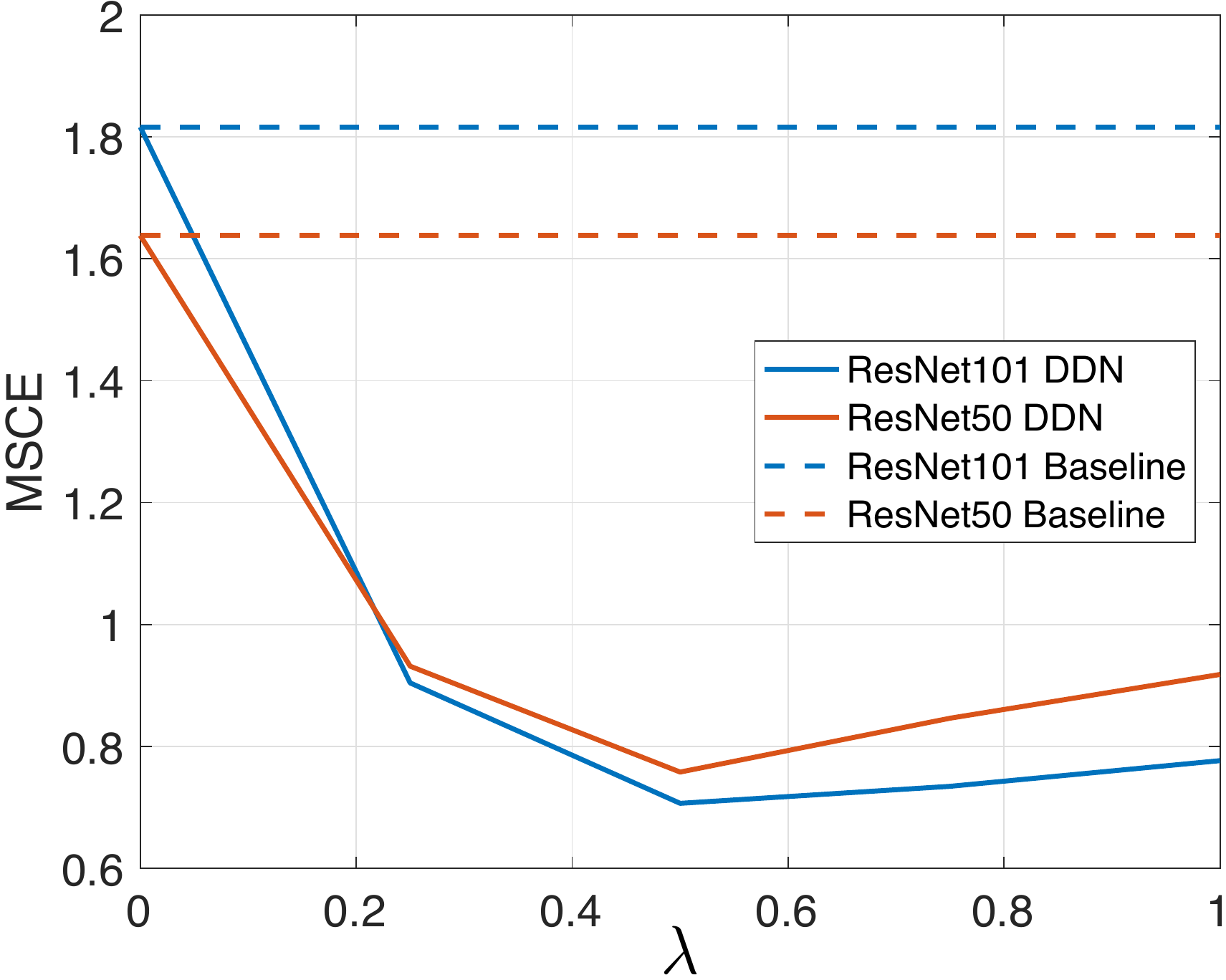}
    \caption{MSCE}
    \label{iou_03_} 
\end{subfigure}
\hspace{5mm}
\begin{subfigure}[]{0.28\textwidth}
    \includegraphics[width=1\textwidth]{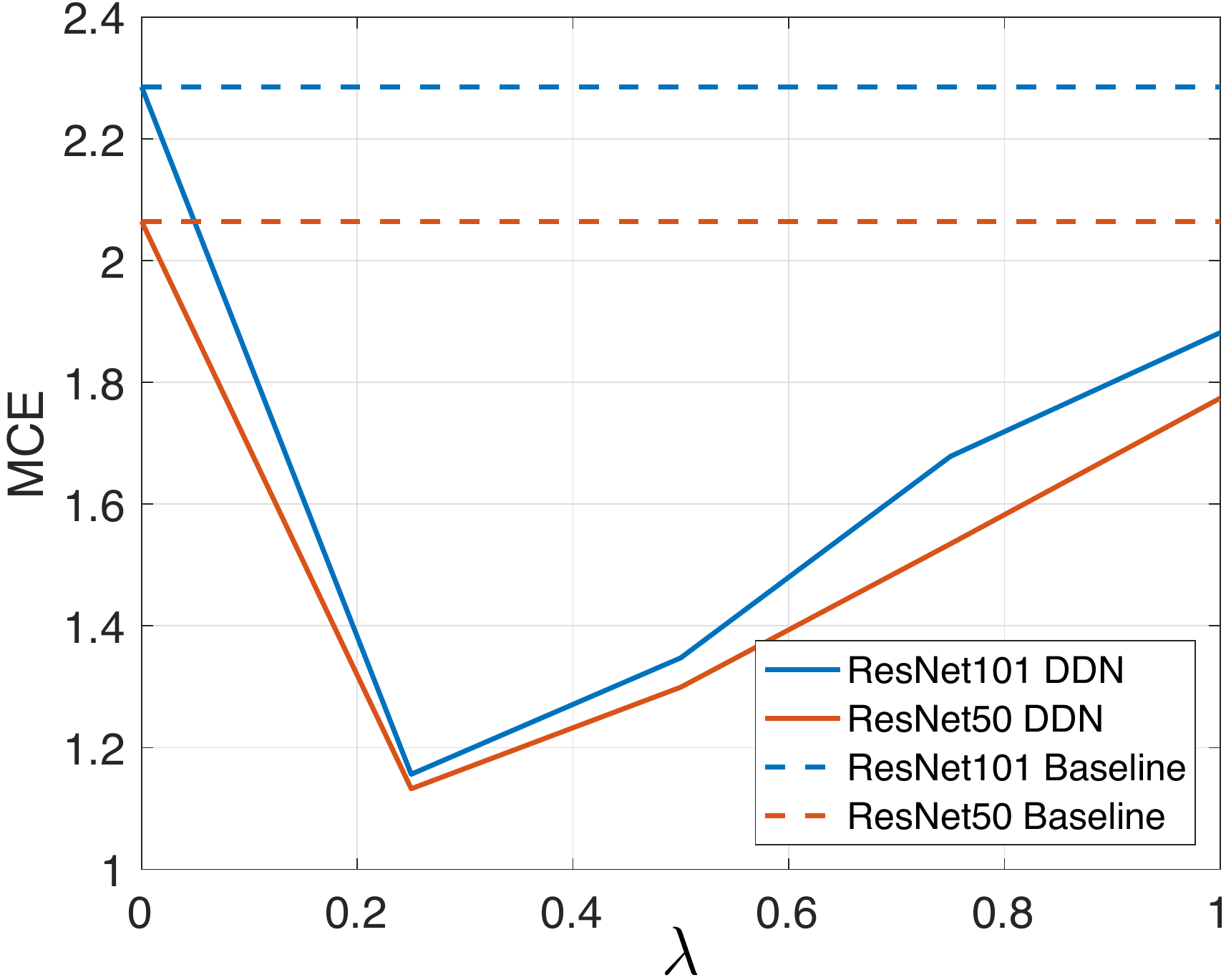}
    \caption{MCE}
    \label{iou_05_}
\end{subfigure}
\hspace{5mm}
\begin{subfigure}[]{0.28\textwidth}
    \includegraphics[width=1\textwidth]{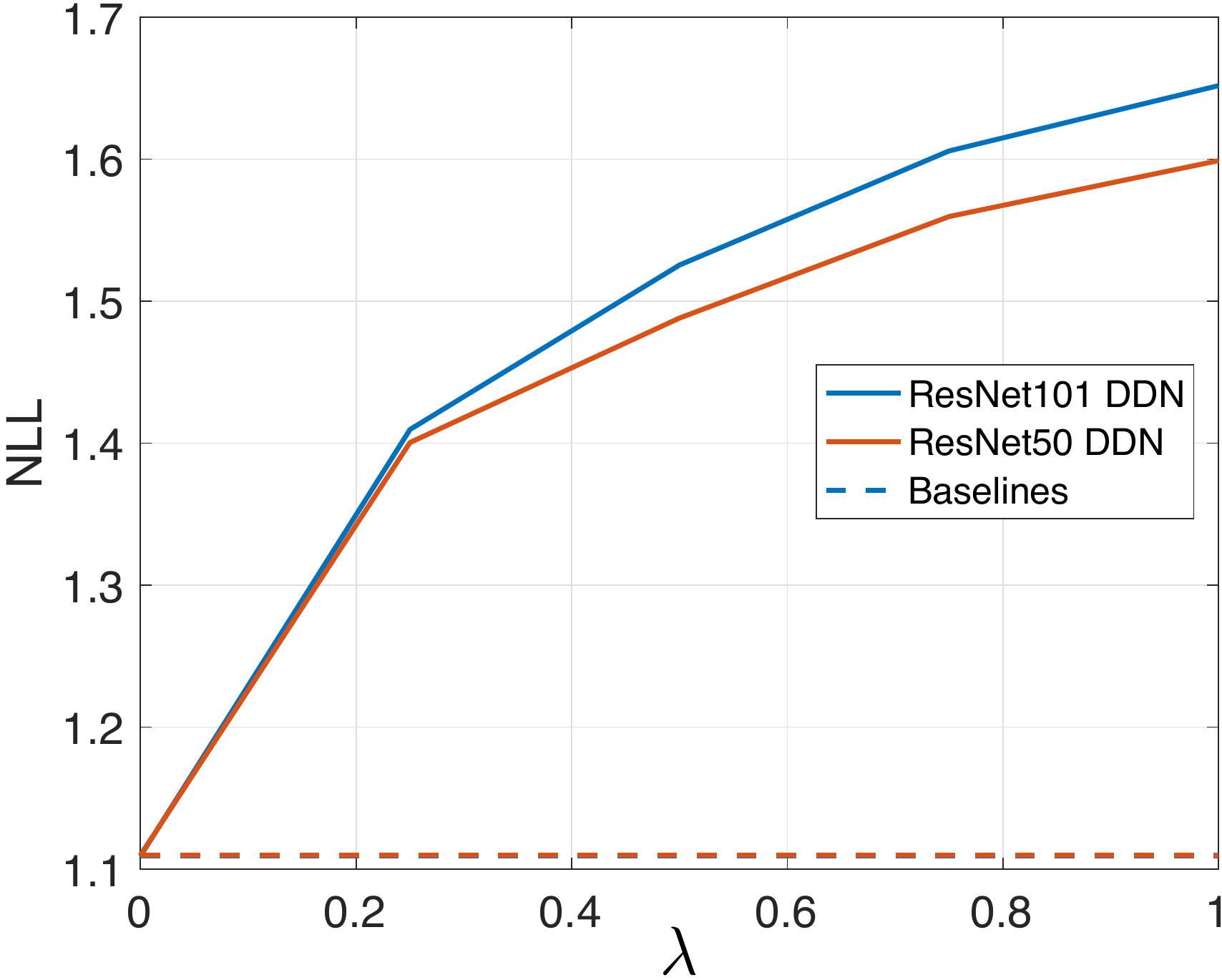}
    \caption{NLL}
    \label{iou_07_}
\end{subfigure}
\caption{Calibration scores for various metrics for different weight values $\lambda$ on the distillation loss. IoU threshold for generating this plot has been set to $0.5$.}
\label{fig:frcnn_weighting}
\end{figure*}

Figure \ref{fig:frcnn_weighting} shows the MSCE and MCE results for different values of the weight parameter $\lambda$ on the distilled dropout loss.
We also show the baseline models which do not use a distillation loss for comparison.
The results in Table \ref{tab:frcnn_calib_res} are reported for $\lambda = 0.5$.

Table~\ref{tab:frcnn_calib_res} shows that at a very small decrease in mean average precision (mAP) and the mean average recall (mAR), for both ResNet50 and ResNet101, DDN shows a significant improvement in MSCE and MCE scores.
Both mAP and mAR are reported for all IoU overlap thresholds between $50\%$ and $95\%$.
The fine-tuning has been performed for an additional $1000$ epochs (more fine-tuning was found to have diminishing returns) on the training and validation dataset.
The evaluation results have been computed on the test dataset consisting of $5000$ images.
While for the classification task DDN shows an improvement in classification accuracy as well as calibration, for object detection it does not appear to make a significant difference.
We believe this is due to the fact that the student model is not trained from scratch and that the training procedure requires additional losses (such as the bounding box regression loss) which also have an influence on the detection accuracy.

\begin{table}[]\centering
\begin{adjustbox}{width=0.5\textwidth}
\begin{tabular}{@{}cccccccccc@{}}
    \midrule
        & & $\mathrm{mAP}_{[.5,.95]}$ & $\mathrm{mAR}_{[.5,.95]}$ & $\mathrm{MSCE}$ & $\mathrm{MCE}$ & $\mathrm{NLL}$ \\
    \midrule
\multirow{2}{*}{\sc{res50}}  & Baseline & \textbf{0.292} & \textbf{0.413} & 1.638 & 2.064 & \textbf{1.109}\\
        & DDN & 0.292 & 0.408 &  \textbf{0.757} & \textbf{1.299} & 1.488\\
    \midrule
    \multirow{2}{*}{\sc{res101}} & Baseline & \textbf{0.320} & \textbf{0.442} & 1.816 & 2.285 & \textbf{1.109}\\
        & DDN & 0.318 & 0.433 & \textbf{0.706} & \textbf{1.347} & 1.525\\
    \bottomrule
\end{tabular}
\end{adjustbox}
\caption[DNN (within Faster R-CNN) calibration and accuracy results on COCO test set]{Faster R-CNN test results for baseline models (ResNet50 and Resnet101) with no dropout and DDN models.
Best outcome is shown in bold.
In terms of calibration (MSCE, MCE), DDN shows a significant improvement over the baseline at a very small cost in terms of model accuracy.}
\label{tab:frcnn_calib_res}
\end{table}

Figure \ref{fig:frcnn_calib_res} shows the calibration improvement of DDN models over the baseline Faster R-CNN models for both ResNet50 and Resnet101.
Subfigures \ref{iou_03}, \ref{iou_05} and \ref{iou_07} correspond to different IoU overlap thresholds of $0.3$, $0.5$ \resp $0.7$ used to distinguish between true and false positive detections, while the IoU threshold for obtaining the target scores for distillation which has been set to $0.5$.
The outcome of using the average of $10$ Monte Carlo samples (MC-10) is also shown for comparison.
On the object detection task, the calibration outcome (measured by MSCE) of our proposed model is less close to that of the ensemble when compared to the calibration results on the classification task.
This could be due to the fact that only $10$ MC-dropout samples have been used to generate the distillation training targets, while for classification the targets were generated from $100$ samples.

\section{Discussion}
A prediction made by sampling the model space or by performing full Bayesian inference takes into account the uncertainty in model parameters $w$. 
The advantage that Bayesian models have over point-estimates of the posterior predictive distribution is that they incorporate this uncertainty into the prediction.
The model that we propose is still a point estimate that only offers an approximation to multiple dropout samples from the model space and it might not be able to capture as well as a Bayesian model when a test sample should have high uncertainty. 
It remains to be investigated how the two types of inference compare when the test data is very different from the training data.
Future work consists of investigating the performance of DDN on out-of-distribution and adversatial input samples. 

\section{Conclusion}
In this work we introduce an efficient way to output better calibrated uncertainty scores from neural networks used for classification and detection.
The Distilled Dropout Network (DDN) makes standard neural networks more introspective and more readily suitable to be deployed in real world applications, where multiple Monte Carlo samples can incur a significant slowdown depending on the number of samples required.
DDN is trained with knowledge distilled from a teacher model using an additional loss which prevents it from being overconfident. 
We demonstrate on the tasks of image classification and image detection that our proposed method improves model calibration and confirm the hypothesis that student models increase task accuracy over their teachers.

\appendix
\section{Qualitative Examples}

\begin{figure*}
    \centering
    \begin{subfigure}[b]{1\textwidth}
        \vspace{4mm} \centering
        \includegraphics[width=0.7\textwidth]{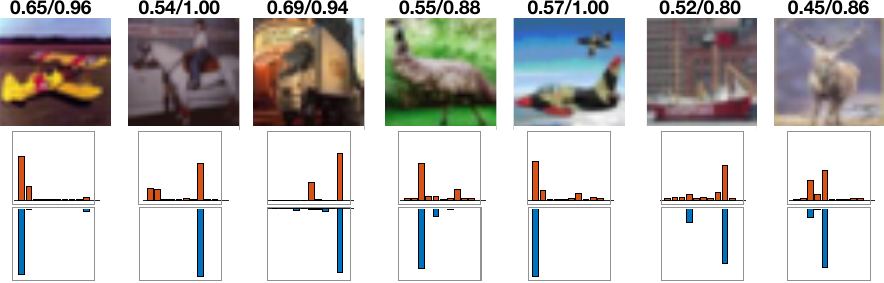} 
        \caption{
        Test samples whose classification outcome is correct for both the baseline model and for DDN. 
        The samples chosen illustrate the most prevalent outcome scenario in which the output distribution predicted by DDN has higher entropy than the one predicted by the baseline.}
        \label{correct_correct_higher_entropy} 
    \end{subfigure}
    
    \begin{subfigure}[b]{1\textwidth}
        \vspace{4mm} \centering
        \includegraphics[width=0.7\textwidth]{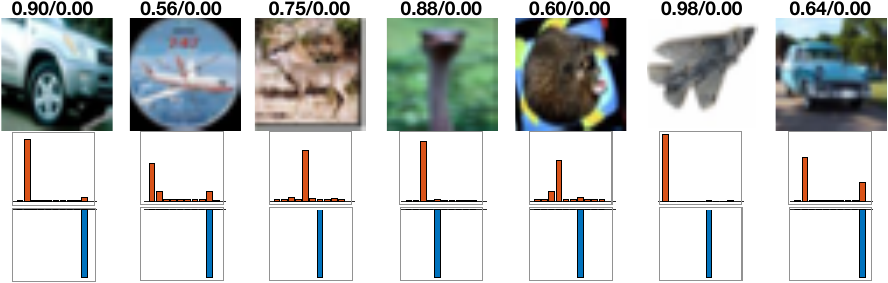} 
        \caption{Test samples whose classification outcome is incorrect for the baseline model but correct for DDN. Moreover, the baseline model makes its predictions with extremely high confidence such that the true class gets assigned almost 0 probability.}
        \label{ddn_wrong_to_correct_highprob_mc1} 
    \end{subfigure}
    
    \begin{subfigure}[b]{1\textwidth}
        \vspace{4mm} \centering
        \includegraphics[width=0.7\textwidth]{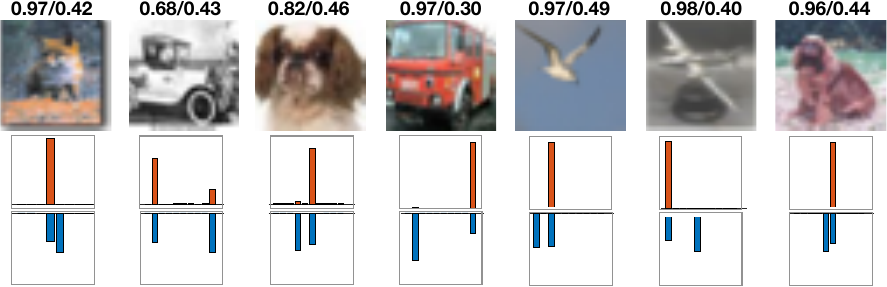} 
        \caption{
        Test samples whose classification outcome is incorrect for the baseline model but correct for DDN. Similar to the previous examples, they illustrate the success cases of our proposed method, in which it doesn't only improve calibration, but it also increases the accuracy of the model.}
        \label{ddn_wrong_to_correct} 
    \end{subfigure}

    \begin{subfigure}[b]{1\textwidth}
        \vspace{4mm} \centering
        \includegraphics[width=0.7\textwidth]{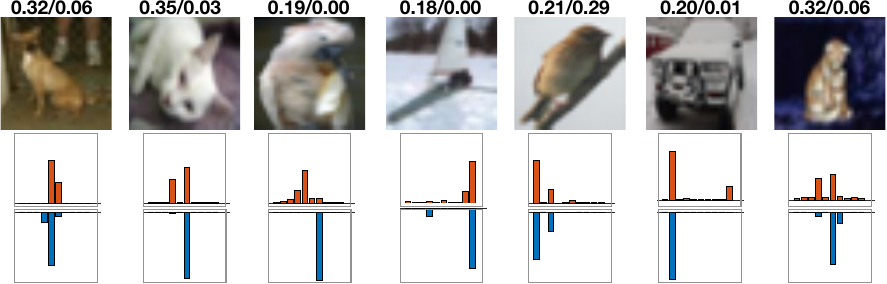} 
        \caption{Test samples incorrectly classified by both models to which DDN assigns a lower probability to the top scoring (wrong) class. 
        In almost all cases, DDN assigns some probability to the true class as well, often assigning to it its second highest probability score.
        Making mistakes with lower confidence is one of the essential characteristics of an introspective classifier.
        }
        \label{ddn_wrong_to_wrong_more_uncert} 
    \end{subfigure}
\caption[Qualitative examples of DDN on the CIFAR-10 test set]{Qualitative examples of CIFAR-10 classification. 
The probability assigned to the true class is shown on top of each image sample, with the score assigned by DDN on the left and the score assigned by the baseline model on the right.
Below the test sample, we show the probability scores assigned to all $10$ output classes by DDN (in red) and by the baseline (in blue, as an inverted histogram).
\label{fig:qualit_cifar_1}
}
\end{figure*}

Figure~\ref{fig:qualit_cifar_1} shows some examples of confidence scores assigned by DDN as well as the baseline on the CIFAR-10 test set.
Figure~\ref{correct_correct_higher_entropy} shows examples in which both models score the true class highest but DDN also assigns some probability to similar output classes.
Such examples are most prevalent and demonstrate DDN's capacity to be more uncertain in its predictions.
Figures~\ref{ddn_wrong_to_correct_highprob_mc1} and \ref{ddn_wrong_to_correct} show some of the samples for which the baseline model assigns the highest probability to an incorrect class while DDN classified the sample correctly.
Such examples make up for $6.75\%$ of the test dataset.
They are the success cases of DDN as the student network is able to change the prediction outcome and increase the task accuracy in addition to improving the model calibration.
We note that the opposite, \ie the percentage of samples classified correctly by the baseline but not by DNN, only makes up for $3.45\%$ of the test dataset.
In the majority of the success case examples the increase in probability assigned to the true class between the baseline and DDN is substantial (the mean is $0.64$ and the variance is $0.04$).
Figure~\ref{ddn_wrong_to_correct_highprob_mc1} shows examples to which the baseline model assigns a confidence score higher than $0.99$ to an incorrect class (overconfident and wrong).
Figure~\ref{ddn_wrong_to_correct} shows some examples for which the baseline model was close to correctly classifying the sample and its mistakes are more understandable (\ie the bird image gets classified as a plane, the dog as a cat or the car as a truck).
A percentage of $6.56\%$ of test samples are classified incorrectly by both models.
On $70\%$ of these examples, DDN assigns a higher probability to the true class.
Figure~\ref{ddn_wrong_to_wrong_more_uncert} shows some of these examples.
In fact, $83.23\%$ of all test samples are in this category.

\begin{figure*}
\centering

    \begin{subfigure}[b]{0.75\textwidth}
        \centering
        \includegraphics[width=1\textwidth]{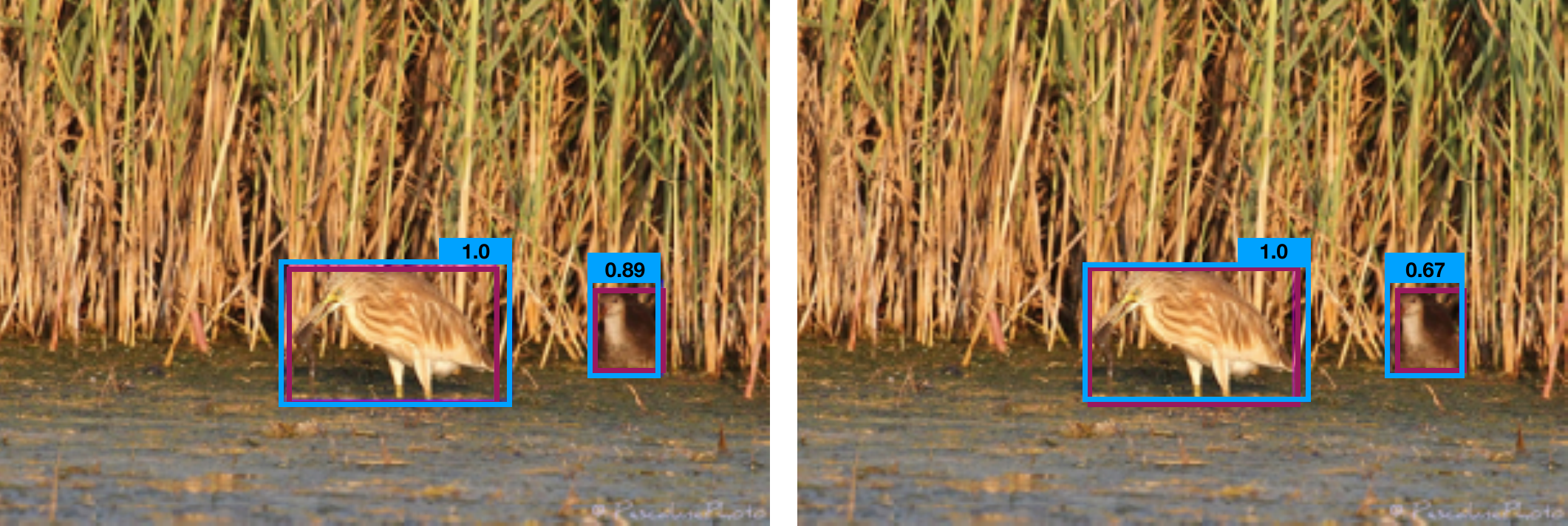} 
        \caption{Example detections to which DDN (right) assigns a probability of $1.0$, same as the baseline (left) to the larger, centred object, while decreasing the confidence assigned to the smaller detection box which seems visually more difficult to classify correctly. In this case, the box whose confidence is decreased is a true positive detection.}
    \end{subfigure}
    \vspace{3mm}
    \begin{subfigure}[b]{0.75\textwidth}
        \centering
        \includegraphics[width=1\textwidth]{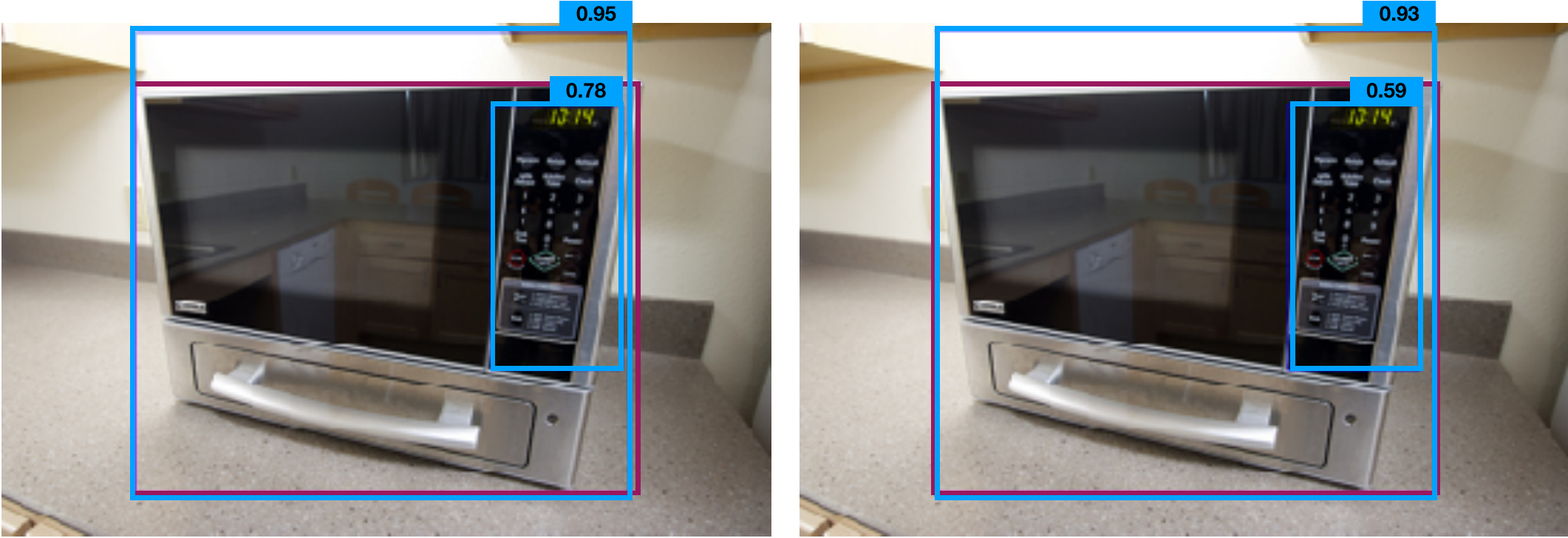} 
        \caption{Example detections to which both DDN and the baseline assign a high confidence to the large, centered object. 
        DDN assigns a lower confidence to the false positive detection, which is the expected behaviour for the proposed model.
          }
    \end{subfigure}
    \vspace{3mm}    
    \begin{subfigure}[b]{0.75\textwidth}
        \centering
        \includegraphics[width=1\textwidth]{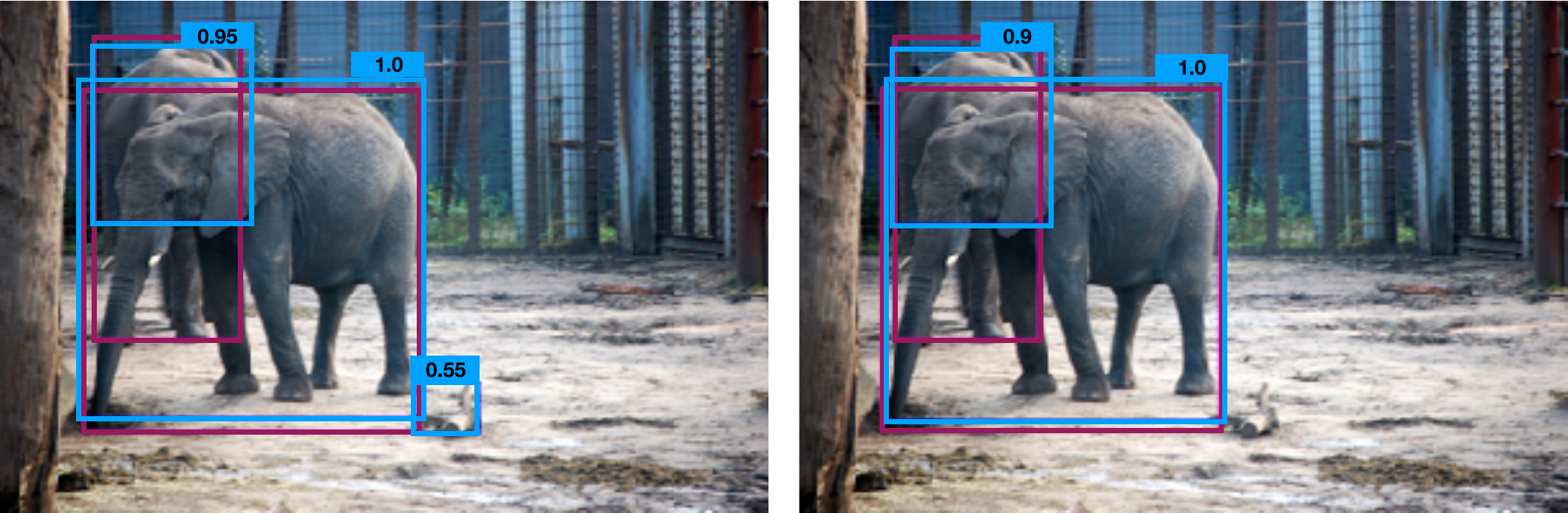} 
        \caption{Example image in which both models assign a high probability to the true positive detections. Furthermore DDN reduces the confidence assigned to the false positive prediction such that it corrects a detector operating with a threshold of $0.5$ on the detection scores.}
    \end{subfigure}
    
\caption[Qualitative examples for DDN on the COCO dataset]{Faster R-CNN detection examples shown for a baseline model (on the left) and DNN (on the right). The magenta bounding boxes are the ground truth annotations and the blue bounding boxes are the predicted detections shown together with the class confidence score. The detection model used is ResNet50 operating with a detection threshold of $0.5$. }
\label{fig:frcnn_qualit}
\end{figure*}

Figure~\ref{fig:frcnn_qualit} shows some example of Faster R-CNN bounding box predictions using the baseline and proposed models as well as their associated ground truth bounding boxes. 
Despite the region proposals being fixed, the predicted bounding boxes are not identical due to the bounding box regression procedure and non-maximum suppression.
We show success examples where DDN lowers the confidences of false positive detections when compared with the baseline model as well as failure cases where this score is increased.

\begin{figure*}
\centering
    \begin{subfigure}[b]{0.75\textwidth}
        \centering
        \includegraphics[width=1\textwidth]{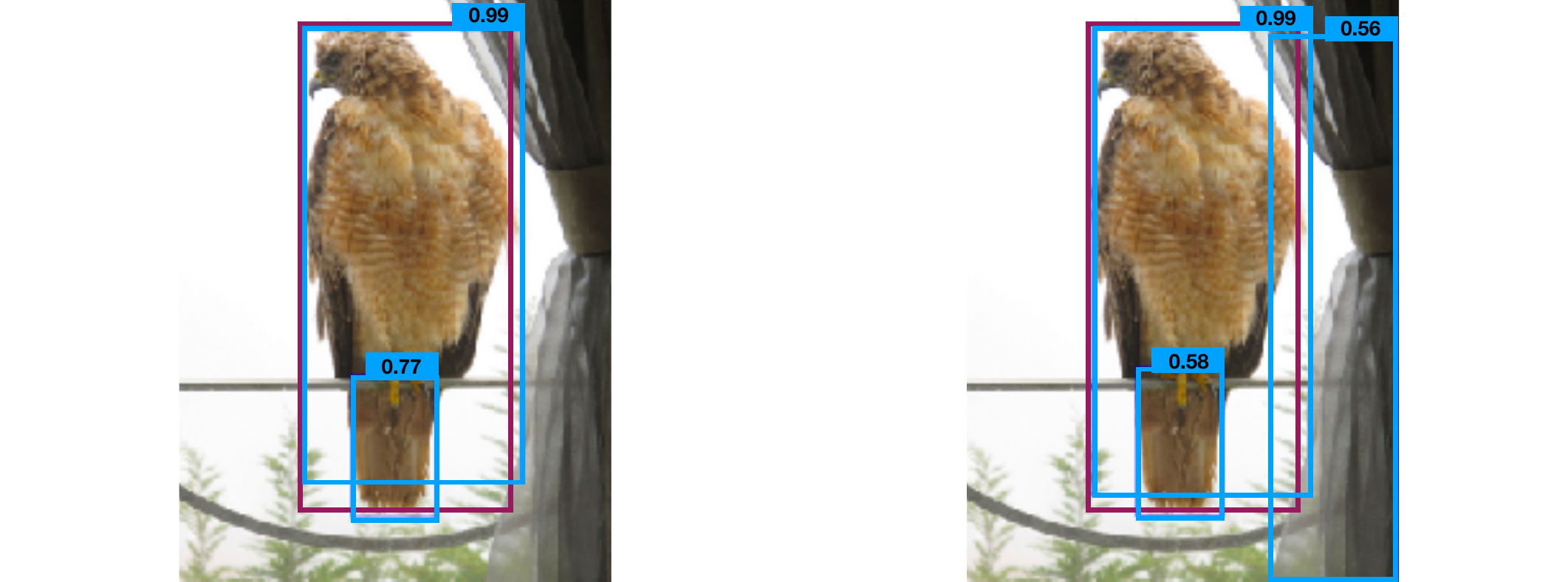} 
        \caption{Example image where DDN assigns a lower confidence score to a false positive detection (the bird feathers) but increases the confidence of another one (the curtain). The latter shows a failure case of our proposed model.} 
        \label{aa}
    \end{subfigure}
    \vspace{3mm}        
    \begin{subfigure}[b]{0.75\textwidth}
        \centering
        \includegraphics[width=1\textwidth]{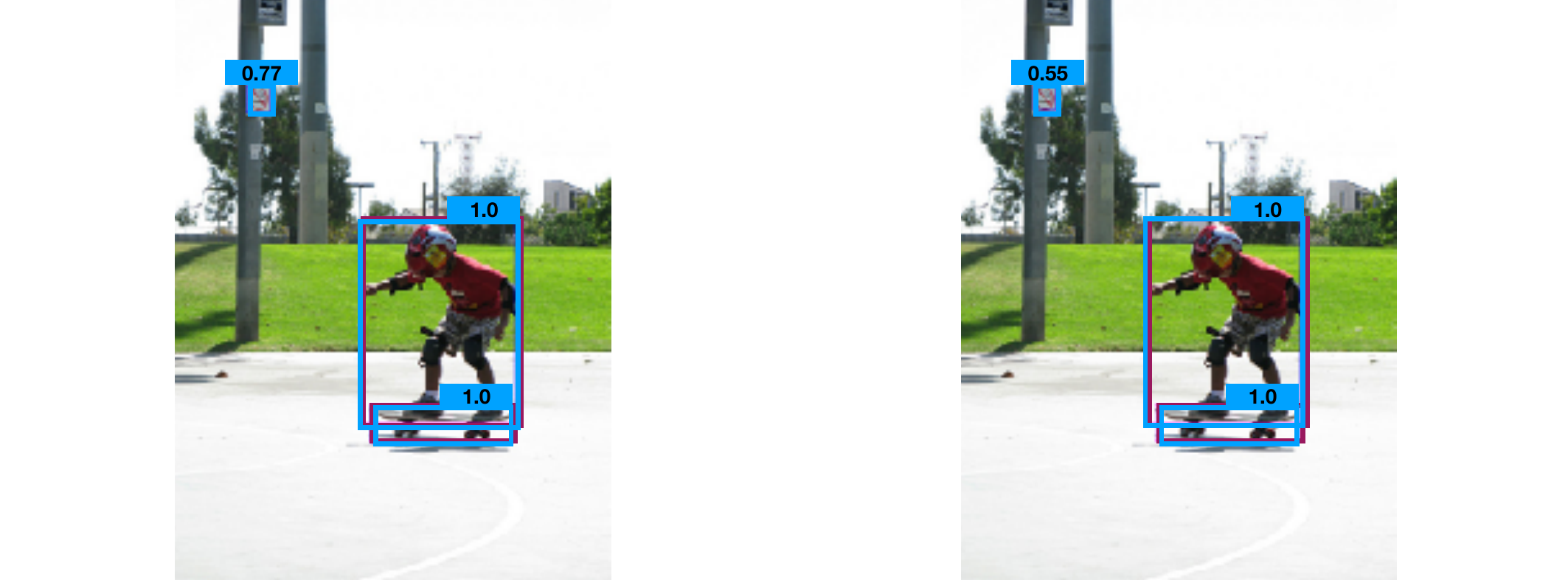} 
        \caption{Example image of DDN decreasing its confidence on a false positive detection (the no smoking sign) while maintaining maximum probability to the true positive detections (child and skateboard).}
        \label{bb}
    \end{subfigure}
    \vspace{3mm}    
    \begin{subfigure}[b]{0.75\textwidth}
        \centering
        \includegraphics[width=1\textwidth]{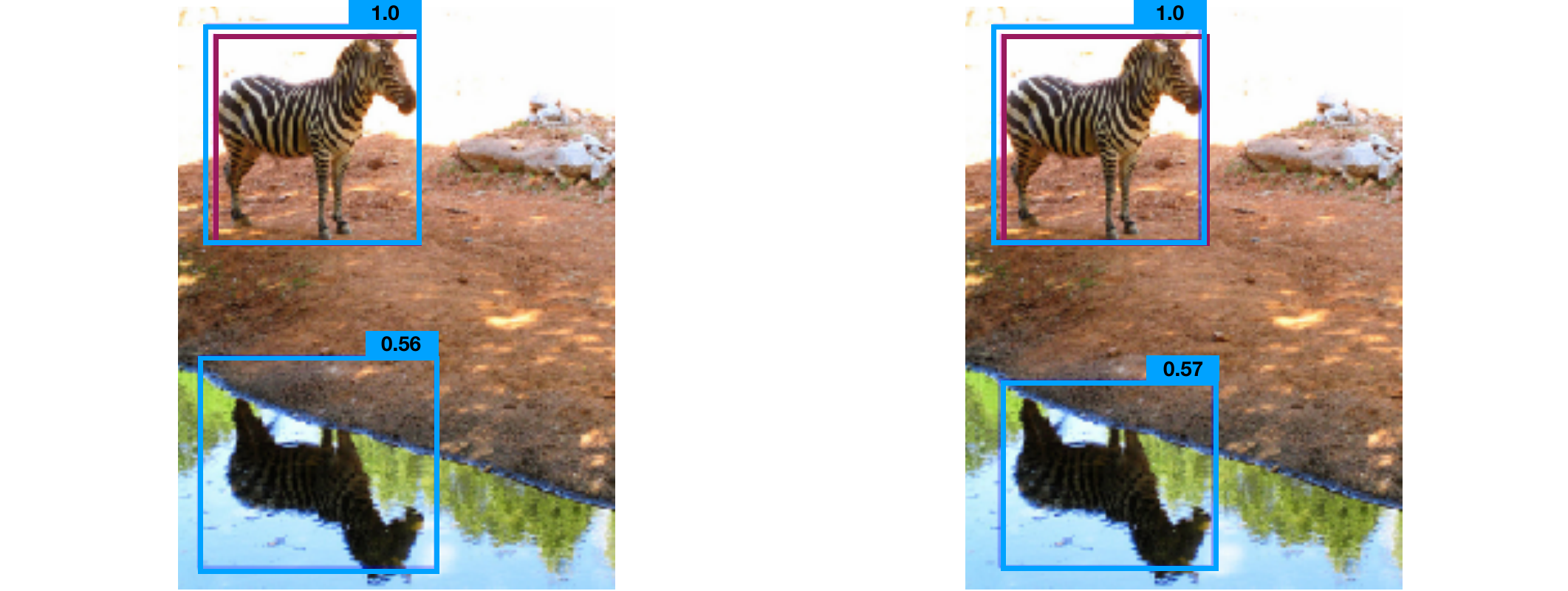} 
        \caption{Example image where DDN increases its confidence on a false positive detection (failure case). This example shows an interesting classification case where the object is visually very similar to the true classification class.}
        \label{cc}
    \end{subfigure}
        
\caption[Qualitative examples for DDN on the COCO dataset]{Faster R-CNN detection examples shown for a baseline model (on the left) and DNN (on the right). The magenta bounding boxes are the ground truth annotations and the blue bounding boxes are the predicted detections shown together with the class confidence score. The detection model used is ResNet50 operating with a detection threshold of $0.5$. }
\label{fig:frcnn_qualit_2}
\end{figure*}

\bibliographystyle{IEEEtran}
\bibliography{root}

\begin{thebibliography}{10}
\providecommand{\url}[1]{#1}
\csname url@samestyle\endcsname
\providecommand{\newblock}{\relax}
\providecommand{\bibinfo}[2]{#2}
\providecommand{\BIBentrySTDinterwordspacing}{\spaceskip=0pt\relax}
\providecommand{\BIBentryALTinterwordstretchfactor}{4}
\providecommand{\BIBentryALTinterwordspacing}{\spaceskip=\fontdimen2\font plus
\BIBentryALTinterwordstretchfactor\fontdimen3\font minus
  \fontdimen4\font\relax}
\providecommand{\BIBforeignlanguage}[2]{{%
\expandafter\ifx\csname l@#1\endcsname\relax
\typeout{** WARNING: IEEEtran.bst: No hyphenation pattern has been}%
\typeout{** loaded for the language `#1'. Using the pattern for}%
\typeout{** the default language instead.}%
\else
\language=\csname l@#1\endcsname
\fi
#2}}
\providecommand{\BIBdecl}{\relax}
\BIBdecl

\bibitem{Blundell_weight_uncert}
\BIBentryALTinterwordspacing
C.~Blundell, J.~Cornebise, K.~Kavukcuoglu, and D.~Wierstra, ``Weight
  uncertainty in neural networks,'' in \emph{Proceedings of the 32Nd
  International Conference on International Conference on Machine Learning -
  Volume 37}, ser. ICML'15.\hskip 1em plus 0.5em minus 0.4em\relax JMLR.org,
  2015, pp. 1613--1622. [Online]. Available:
  \url{http://dl.acm.org/citation.cfm?id=3045118.3045290}
\BIBentrySTDinterwordspacing

\bibitem{dropout}
\BIBentryALTinterwordspacing
N.~Srivastava, G.~Hinton, A.~Krizhevsky, I.~Sutskever, and R.~Salakhutdinov,
  ``Dropout: A simple way to prevent neural networks from overfitting,''
  \emph{Journal of Machine Learning Research}, vol.~15, pp. 1929--1958, 2014.
  [Online]. Available: \url{http://jmlr.org/papers/v15/srivastava14a.html}
\BIBentrySTDinterwordspacing

\bibitem{Gal2015DropoutB}
Y.~Gal and Z.~Ghahramani, ``Dropout as a {B}ayesian approximation: Representing
  model uncertainty in deep learning,'' \emph{arXiv:1506.02142}, 2015.

\bibitem{Gal2016Uncertainty}
Y.~Gal, ``Uncertainty in deep learning,'' Ph.D. dissertation, University of
  Cambridge, 2016.

\bibitem{simple_and_scalable}
B.~Lakshminarayanan, A.~Pritzel, and C.~Blundell, ``Simple and scalable
  predictive uncertainty estimation using deep ensembles,'' 12 2016.

\bibitem{ensemble_methods_in_ml}
T.~G. Dietterich, ``Ensemble methods in machine learning,'' in \emph{Multiple
  Classifier Systems}.\hskip 1em plus 0.5em minus 0.4em\relax Berlin,
  Heidelberg: Springer Berlin Heidelberg, 2000, pp. 1--15.

\bibitem{GrimmettIJRR2015}
H.~Grimmett, R.~Triebel, R.~Paul, and I.~Posner, ``Introspective classification
  for robot perception,'' \emph{The International Journal of Robotics
  Research}, vol.~35, no.~7, pp. 743--762, 2016.

\bibitem{regularising_hinton}
\BIBentryALTinterwordspacing
G.~Pereyra, G.~Tucker, J.~Chorowski, L.~Kaiser, and G.~E. Hinton,
  ``Regularizing neural networks by penalizing confident output
  distributions,'' \emph{CoRR}, vol. abs/1701.06548, 2017. [Online]. Available:
  \url{http://arxiv.org/abs/1701.06548}
\BIBentrySTDinterwordspacing

\bibitem{BucilaCN06}
\BIBentryALTinterwordspacing
C.~Bucila, R.~Caruana, and A.~Niculescu{-}Mizil, ``Model compression,'' in
  \emph{Proceedings of the Twelfth {ACM} {SIGKDD} International Conference on
  Knowledge Discovery and Data Mining, Philadelphia, PA, USA, August 20-23,
  2006}, 2006, pp. 535--541. [Online]. Available:
  \url{http://doi.acm.org/10.1145/1150402.1150464}
\BIBentrySTDinterwordspacing

\bibitem{BANNs}
T.~Furlanello, Z.~C. Lipton, M.~Tschannen, L.~Itti, and A.~Anandkumar, ``Born
  again neural networks,'' 2017.

\bibitem{Bulo_dropout_distill}
\BIBentryALTinterwordspacing
S.~R. Bul\`{o}, L.~Porzi, and P.~Kontschieder, ``Dropout distillation,'' in
  \emph{Proceedings of the 33rd International Conference on International
  Conference on Machine Learning - Volume 48}, ser. ICML'16.\hskip 1em plus
  0.5em minus 0.4em\relax JMLR.org, 2016, pp. 99--107. [Online]. Available:
  \url{http://dl.acm.org/citation.cfm?id=3045390.3045402}
\BIBentrySTDinterwordspacing

\bibitem{on_calibr_on_nn}
\BIBentryALTinterwordspacing
C.~Guo, G.~Pleiss, Y.~Sun, and K.~Q. Weinberger, ``On calibration of modern
  neural networks,'' \emph{CoRR}, vol. abs/1706.04599, 2017. [Online].
  Available: \url{http://arxiv.org/abs/1706.04599}
\BIBentrySTDinterwordspacing

\bibitem{distilling}
\BIBentryALTinterwordspacing
G.~Hinton, O.~Vinyals, and J.~Dean, ``Distilling the knowledge in a neural
  network,'' in \emph{NIPS Deep Learning and Representation Learning Workshop},
  2015. [Online]. Available: \url{http://arxiv.org/abs/1503.02531}
\BIBentrySTDinterwordspacing

\bibitem{kendall2017uncertainties}
A.~Kendall and Y.~Gal, ``What uncertainties do we need in bayesian deep
  learning for computer vision?'' \emph{arXiv preprint arXiv:1703.04977}, 2017.

\bibitem{NIPS2014_5484}
\BIBentryALTinterwordspacing
J.~Ba and R.~Caruana, ``Do deep nets really need to be deep?'' in
  \emph{Advances in Neural Information Processing Systems 27}, Z.~Ghahramani,
  M.~Welling, C.~Cortes, N.~D. Lawrence, and K.~Q. Weinberger, Eds.\hskip 1em
  plus 0.5em minus 0.4em\relax Curran Associates, Inc., 2014, pp. 2654--2662.
  [Online]. Available:
  \url{http://papers.nips.cc/paper/5484-do-deep-nets-really-need-to-be-deep.pdf}
\BIBentrySTDinterwordspacing

\bibitem{degroot}
M.~H. DeGroot and S.~E. Fienberg, ``The comparison and evaluation of
  forecasters,'' \emph{The Statistician: Journal of the Institute of
  Statisticians}, vol.~32, pp. 12--22, 1983.

\bibitem{Gal2017Concrete}
Y.~Gal, J.~Hron, and A.~Kendall, ``{Concrete Dropout},'' in
  \emph{arXiv:1705.07832}, May 2017.

\bibitem{OndruskaIVS2014}
P.~Ondruska and I.~Posner, ``Probabilistic attainability maps: Efficiently
  predicting driver-specific electric vehicle range,'' in \emph{IEEE
  Intelligent Vehicles Symposium (IV)}, Dearborn, MI, USA, June 2014.

\bibitem{Krizhevsky09learningmultiple}
A.~Krizhevsky, ``Learning multiple layers of features from tiny images,'' Tech.
  Rep., 2009.

\bibitem{huang2017densely}
G.~Huang, Z.~Liu, L.~van~der Maaten, and K.~Q. Weinberger, ``Densely connected
  convolutional networks,'' in \emph{Proceedings of the IEEE Conference on
  Computer Vision and Pattern Recognition}, 2017.

\bibitem{pmlr-v28-sutskever13}
\BIBentryALTinterwordspacing
I.~Sutskever, J.~Martens, G.~Dahl, and G.~Hinton, ``On the importance of
  initialization and momentum in deep learning,'' in \emph{Proceedings of the
  30th International Conference on Machine Learning}, ser. Proceedings of
  Machine Learning Research, S.~Dasgupta and D.~McAllester, Eds., vol.~28,
  no.~3.\hskip 1em plus 0.5em minus 0.4em\relax Atlanta, Georgia, USA: PMLR,
  17--19 Jun 2013, pp. 1139--1147. [Online]. Available:
  \url{http://proceedings.mlr.press/v28/sutskever13.html}
\BIBentrySTDinterwordspacing

\bibitem{ren2015faster}
\BIBentryALTinterwordspacing
S.~Ren, K.~He, R.~B. Girshick, and J.~Sun, ``Faster {R-CNN:} towards real-time
  object detection with region proposal networks,'' \emph{CoRR}, vol.
  abs/1506.01497, 2015. [Online]. Available:
  \url{http://arxiv.org/abs/1506.01497}
\BIBentrySTDinterwordspacing

\bibitem{mscoco}
T.-Y. Lin, M.~Maire, S.~Belongie, J.~Hays, P.~Perona, D.~Ramanan,
  P.~Doll{\'a}r, and C.~L. Zitnick, ``Microsoft coco: Common objects in
  context,'' in \emph{Computer Vision -- ECCV 2014}, D.~Fleet, T.~Pajdla,
  B.~Schiele, and T.~Tuytelaars, Eds.\hskip 1em plus 0.5em minus 0.4em\relax
  Cham: Springer International Publishing, 2014, pp. 740--755.

\bibitem{He2015}
K.~He, X.~Zhang, S.~Ren, and J.~Sun, ``Deep residual learning for image
  recognition,'' \emph{arXiv preprint arXiv:1512.03385}, 2015.

\bibitem{chen17implementation}
X.~Chen and A.~Gupta, ``An implementation of faster rcnn with study for region
  sampling,'' \emph{arXiv preprint arXiv:1702.02138}, 2017.

\bibitem{imagenet}
J.~Deng, W.~Dong, R.~Socher, L.-J. Li, K.~Li, and L.~Fei-Fei, ``{ImageNet: A
  Large-Scale Hierarchical Image Database},'' in \emph{CVPR09}, 2009.

\end{thebibliography}

\end{document}